\documentclass[11pt,a4paper]{article}
\usepackage[hyperref]{ranlp2021}
\usepackage{times}
\usepackage{latexsym}

\usepackage{multirow}
\usepackage{xcolor}
\usepackage{booktabs}

\usepackage{microtype}

\aclfinalcopy % Uncomment this line for the final submission

\title{Generating Answer Candidates\\ for Quizzes and Answer-Aware Question Generators}

\author{
Kristiyan Vachev \\
FMI, Sofia University \\
``St. Kliment Ohridski'' \\
Sofia, Bulgaria\\
{\small \texttt{vachevkd@gmail.com}}
  \And
  Momchil Hardalov \\
  FMI, Sofia University \\ ``St. Kliment Ohridski'' \\
  Sofia, Bulgaria\\
{\small \texttt{hardalov@fmi.uni-sofia.bg}}
 \\\And
  Georgi Karadzhov \\
  Department of Computer \\ Science and Technology\\
  University of Cambridge, UK \\
  {\small \texttt{gmk34@cam.ac.uk}}
   \AND
  Georgi Georgiev \\ 
  Releva AI \\
  Sofia, Bulgaria \\
  {\small \texttt{georgi@releva.ai}}
  \And
  Ivan Koychev \\
  FMI and GATE\\
  Sofia University \\
  Sofia, Bulgaria\\
 {\small \texttt{koychev@fmi.uni-sofia.bg}}
  \\\And
  Preslav Nakov \\
  Qatar Computing Research Institute\\ HBKU\\
  Doha, Qatar\\
  {\small \texttt{pnakov@hbku.edu.qa}}
}

\date{}

\usepackage{graphicx}
\graphicspath{ {./images/} }

\begin{document}
\maketitle

\begin{abstract}
In education, open-ended quiz questions have become an important tool for assessing the knowledge of students. Yet, manually preparing such questions is a tedious task, and thus automatic question generation has been proposed as a possible alternative. So far, the vast majority of research has focused on generating the question text, relying on question answering datasets with readily picked answers, and the problem of how to come up with answer candidates in the first place has been largely ignored. Here, we aim to bridge this gap. In particular, we propose a model that can generate a specified number of answer candidates for a given passage of text, which can then be used by instructors to write questions manually or can be passed as an input to automatic answer-aware question generators. Our experiments show that our proposed answer candidate generation model outperforms several baselines.
\end{abstract}

\begin{figure}[t!]
\textbf{Context:}
\textit{One of the most famous people born in Warsaw was \textbf{Maria Skłodowska-Curie}, who achieved international recognition for her research on radioactivity and was the first female recipient of the \textbf{Nobel Prize}. \textbf{Famous musicians} include Władysław Szpilman and Frédéric Chopin. Though Chopin was born in the village of Żelazowa Wola, about 60 km (37 mi) from Warsaw, he moved to the city with his family when he was \textbf{seven months old}. Casimir Pulaski, a Polish general and hero of the American Revolutionary War, was born here in \textbf{1745}.
} \\

\textbf{Q1}: \textit{What was Maria Curie the first female recipient of?} \\
\textbf{A1}: \textit{Nobel Prize} \\ \\
\textbf{Q2}:\textit{ What year was Casimir Pulaski born in Warsaw?} \\
\textbf{A2}: \textit{1745} \\ \\
\textbf{Q3}: \textit{How old was Chopin when he moved to Warsaw with his family?} \\
\textbf{A3}: \textit{seven months old}

\caption{Example passage from the SQuAD1.1 dataset showing three of the questions created by humans and all answers to those and other questions in bold.}
\label{fig:x sample squad}
\end{figure}

\section{Introduction}

Testing with open-ended quiz questions can help both learning and retention, e.g.,~it could be used for self-study or as a way to detect knowledge gaps in a classroom setting, thus allowing instructors to adapt their teaching~\citep{ROEDIGERIII20111}.

As creating such quiz questions is a tedious job, automatic methods have been proposed. The task is often formulated as an answer-aware question generation \citep{heilman-smith-2010-good,10.1145/2661829.2661908,du2017learning,du-cardie-2018-harvesting,sun-etal-2018-answer,dong2019unified,bao2020unilmv2,8585151}: given an input text and a target answer, generate a corresponding question.

Many researchers have used the Stanford Question Answering Dataset (SQuAD1.1) dataset~\citep{rajpurkar2016squad} as a source of training and testing data for answer-aware question generation. It contains human-generated questions and answers about articles in Wikipedia, as shown in Figure~\ref{fig:x sample squad}. 

However, this formulation requires that answers be picked beforehand, which may not be practical for real-world situations. Here we aim to address this limitation by proposing a method for generating answers, which can in turn serve as an input to answer-aware question generation models. Our model combines orthographic, lexical, syntactic, and semantic information, and shows promising results. It further allows the user to specify the number of answer to propose. Our contributions can be summarized as follows: 

\begin{itemize}
    \item We propose a new task: generate answer candidates that can serve as an input to answer-aware question generation models.
    \item We create a dataset for this new task.
    \item We propose a suitable model for the task, which combines orthographic, lexical, syntactic, and semantic information, and can generate a pre-specified number of answers.
    \item We demonstrate improvements over simple approaches based on named entities, and competitiveness over complex neural models.
\end{itemize}

\section{Related Work}

The success of large-scale pre-trained Transformers such as BERT~\citep{devlin2019bert}, RoBERTa~\citep{liu2019roberta}, ALBERT~\citep{lan2020albert}, and generative ones such as T5~\citep{raffel2020exploring} or BART~\citep{lewis2019bart}, has led to the rise in popularity of the Question Generation task. Models such as BERT~\citep{devlin2019bert}, T5~\citep{raffel2020exploring} and PEGASUS~\citep{zhang2020pegasus} have been used to generate questions for the SQuAD1.1 dataset and have been commonly evaluated \citep{du2017learning} using BLEU~\citep{papineli2002bleu}, ROUGE~\citep{lin2004rouge}, and METEOR~\citep{lavie-agarwal-2007-meteor}. Strong models for this task include NQG++~\citep{zhou2017neural}, ProphetNet~\citep{qi2020prophetnet}, MPQG~\citep{song-etal-2018-leveraging}, UniLM~\citep{dong2019unified}, UniLMv2 \citep{bao2020unilmv2}, and ERNIE-GEN \citep{xiao2020erniegen}.

All these models were trained for answer-aware question generation, which takes the answer and the textual context as an input and outputs a question for that answer. In contrast, our task formulation takes a textual context as an input and generates possible answers; in turn, these answers can be used as an input to the above answer-aware question generation models.  

The Quiz-Style Question Generation for News Stories task \citep{lelkes2021quizstyle} uses a formulation that asks to generate a single question as well as the corresponding answer, which is to be extracted from the given context.

Follow-up research has tried to avoid the limitation of generating a single question or a single question--answer pair by generating a question for each sentence in the input context or by using all named entities in the context as answer keys \citep{questiongenerator20}.

Finally, there has been a proliferation of educational datasets in recent years \citep{zeng2020survey, dzendzik2021english,rogers2021qa}, which includes Crowdsourcing~\citep{welbl2017crowdsourcing}, ARC~\citep{clark2018think}, OpenBookQA~\citep{mihaylov2018suit}, multiple-choice exams in Bulgarian~\citep{hardalov2019englishonly}, Vietnamese~\citep{9247161}, and EXAMS, which covers 16 different languages~\citep{hardalov2020exams}. Yet, these datasets are not directly applicable for our task as their questions do not expect the answers to be exact matches from the textual context. While there are also span-based extraction datasets such as NewsQA~\citep{trischler-etal-2017-newsqa}, SearchQA~\citep{dunn2017searchqa} and Natural Questions: A Benchmark for Question Answering Research~\citep{kwiatkowski-etal-2019-natural} they contains a mix of long and short spans rather than factoid answers. Thus, we opted to use SQuAD1.1 in our experiments, but focusing on generating answers rather than on questions.

\section{Method}

Given an input textual context, we first extract phrases from it, then we calculate a representation for each phrase, and finally, we predict which phrases are appropriate of being an answer to a quiz question based on these representations.

\subsection{Data}

To train our classifier, we need a labeled dataset of key phrases. In particular, we use SQuAD1.1, which consists of more than 100,000 questions created by humans from Wikipedia articles, and was extensively used for question answering. An example is shown in Figure \ref{fig:x sample squad}. We use version 1.1 of the dataset instead of 2.0~\citep{rajpurkar2018know} because it contains the exact position of the answers in the text, which allows us to easily match them against the candidate phrases. Version 2.0 only adds examples whose answer is not present.

We created a dataset for our task using 87,600 questions from the SQuAD1.1 training set and their associated textual contexts. Because only 33\% of the answers consisted of one word, it is important to also extract phrases longer than a single word. Thus, we also added all named entities; note that they have a variable word length. We further included all noun chunks, which we then extended by combining two or more noun chunks if the only words between them were connectors like \emph{and}, \emph{of}, and \emph{or}. Here is an example of a complex chunk with three pieces: \emph{a Marian place \textbf{of} prayer \textbf{and} reflection}. We considered as positive examples the phrases for which there was a question asked in the SQuAD1.1 dataset, and we considered as negative examples the additional phrases we created.

\subsection{Features}

We extracted the following features, adapted for the use of phrases containing multiple words: 

\textbf{TFIDFArticle}, \textbf{TFIDFParagraph}: The average TF.IDF score for all words in the key phrase, where the Inverse Document Frequency (IDF) is computed from the words in all paragraphs of the article (\textit{TFIDFArticle}) or only from the paragraph of the given key phrase (\textit{TFIDFParagraph}).

\textbf{TitleSimilarity}: The average cosine similarity between the vectors of the words in the key phrase and the article title.

\textbf{POS}, \textbf{TAG}, \textbf{DEP}: The coarse-grained part-of-speech tag (\textit{POS}), the fine-grained part-of-speech tag (\textit{TAG}), and the syntactic dependency relation (\textit{DEP}). If the phrase contains multiple words, we only consider the word with the highest TF.IDF.

\textbf{EntityType}: The named entity type of the phrase if any.

\textbf{IsAlpha}: True if all characters in the phrase are alphabetic. 

\textbf{IsAscii}: True if the phrase consists only of characters contained in the standard ASCII table. 

\textbf{IsDigit}: True if the phrase only contains digits. 

\textbf{IsLower}: True if all words in the phrase are in lowercase. 

\textbf{IsCapital}: True if the first word in the phrase is in uppercase.

\textbf{IsCurrency}: True if some word in the phrase contains a currency symbol, e.g., 
\emph{\$23}.

\textbf{LikeNum}: True if some word in the phrase represents a number, e.g.,~\emph{13.4}, \emph{42}, \emph{twenty}, etc.

\subsection{Model}

We convert all the above features to binary, and then we use a Bernoulli Na\"{i}ve Bayes classifier, which can account both for the presence and for the absence of a feature. To achieve this, we encode categorical features (e.g., \emph{POS}, \emph{TAG}) using one-hot encoding, and we put continuous features (e.g., \emph{TFIDFArticle}, \emph{TitleSimilarity}) into bins of five.

\subsection{Evaluation Measures}

As there is no established measure for evaluating key phrases for answer generation, we use and adapt the original evaluation script\footnote{\url{http://github.com/allenai/bi-att-flow/blob/master/squad/evaluate-v1.1.py}} created for the Question Answering task on the SQuAD1.1 dataset \citep{rajpurkar2016squad}, which calculates an exact match (\textit{EM}) and the harmonic mean of precision and recall (\textit{F1}).

In the SQuAD1.1 dataset, there can be multiple correct versions of the answer for a question (e.g.,~\emph{third}, \emph{third-most}). Thus, the evaluation script calculates EM and F1 for each such version and then returns the highest value. As there can be also multiple question--answer pairs in a given passage, we further adapted the script to include all human-created answers, we calculated these scores against all answers in the passage, and finally, we took the highest values. 

Finally, in order to allow for a more practical use of question generation algorithms, it is desirable to be able to generate multiple question--answer pairs for a given passage. To compute EM and F1 over multiple answer candidates, we adopted the following two approaches:

\textbf{EM-Any} and \textbf{F1-Any} show how likely it is to pick a ground-truth answer (also, how likely it is to be chosen by a human annotator of SQuAD1.1) out of \textit{N} returned candidate answers. To calculate them, we took only the best EM and F1 scores after computing all scores for each candidate answer.

Using \textbf{EM-Avg} and \textbf{F1-Avg}, we can measure what percentage of all returned candidate answers have also been marked as an answer by a human. To calculate them, we took the average of all EM and F1 scores computed for the proposed candidate answers. 

The results for the SQuAD1.1 development split, which consists of 2,067 unique passages, are shown in Table \ref{table:1} and Table \ref{table:2}.

\section{Experiments and Evaluation}

We used our model to generate ten candidate answers per passage (taking the ones with the highest classifier confidence), and we compared the results to other commonly used methods.

\subsection{Baselines}
\label{sec:baselines}

Below, we list the baselines that we compared against:

\begin{itemize}
    \item \textbf{NER:} Extracting all named entities from the passage and using them as candidate answers. On average, there are 13.64 named entities per SQuAD1.1 passage.
   
    \item \textbf{Noun Chunks:} Extracting all noun chunks from the passage and using them as candidate answers. On average, there are 33.15 noun chunks per SQuAD1.1 passage.
  
    \item \textbf{NE + NCh:} Combining all extracted named entities and noun chunks from the passage after using the SQuAD1.1 normalization script\footnote{\url{http://github.com/allenai/bi-att-flow/blob/master/squad/evaluate-v1.1.py\#L11}} to remove duplicate words (e.g.,~\emph{the third} matches \emph{third}).
   
    \item \textbf{T5-small:} We fine-tuned the small version of T5, which has 220M parameters. We trained the model to accept the passage as an input and to output the answer. We used a learning rate of 0.0001, a source token length of 300, and a target token length of 24. The best validation loss was achieved in the forth out of ten epochs. 
\end{itemize}

\subsection{Results}

In this section, we describe our experimental results and we compare them to the baselines described in Section~\ref{sec:baselines} above.

\begin{table}[t!]
\centering
\begin{tabular}{lcrr}
\toprule
\textbf{Method} & \textbf{Answers} & \textbf{EM-Any} & \textbf{F1-Any} \\
\midrule
\multirow{10}{*}{Our Model} & 1 & 29.63 & 38.80 \\
 & 2 & 42.50 & 55.47 \\
 & 3 & 52.92 & 66.83 \\
 & 4 & 59.67 & 73.92 \\
 & 5 & 65.50 & 79.11 \\
 & 6 & 69.33 & 82.50 \\
 & 7 & 72.90 & 85.58 \\
 & \bf 8 & \bf 75.43 & \bf 87.70 \\
 & 9 & 77.66 & 89.20 \\
 & 10 & 79.17 & 90.32 \\
\midrule
NER & \bf 13.6 & \bf 74.36 & \bf 82.12 \\
NCh & 33.2 & 86.79 & 95.90 \\
NER + NCh  & 35.4 & 95.02 & 98.48 \\
T5-small & 1 & 37.56 & 49.16 \\
\bottomrule
\end{tabular}
\caption{\textbf{Best over multiple candidates (EM-Any and F1-Any).} Measuring how often, among the top-$N$ candidates proposed by the model, at least one was picked by a human.}
\label{table:1}
\end{table}

\subsubsection{Best Over Multiple Candidates}

Table~\ref{table:1} shows the results for EM-Any and F1-Any, i.e.,~how often, among the top-$N$ candidates by the model, at least one was picked by a human.

We can see that, compared to using named entities, our model achieves a better EM-Any score with just eight answer candidates rather than using all named entities in the passage (which are 13.6 on average). It also achieves a higher F1-Any score with just six answer candidates.

We further see that using the combination of all named entities and noun chunks yields the best score, but it produces 35 candidates on average, which is the majority of the words in the passage.

\begin{table}[t!]
\centering
\begin{tabular}{lcrr}
\toprule
\textbf{Method} & \textbf{Answers} & \textbf{EM-Avg} & \textbf{F1-Avg}\\
\midrule
\multirow{10}{*}{Our Model} & 1 & \bf 29.63 & \bf 38.80 \\
 & 2 & 25.58 & 36.03 \\
 & 3 & 24.18 & 35.15 \\
 & 4 & 22.74 & 34.15 \\
 & 5 & 22.07 & 33.65 \\
 & 6 & 20.90 & 32.81 \\
 & 7 & 20.02 & 32.05 \\
 & 8 & 19.25 & 31.43 \\
 & 9 & 18.45 & 30.57 \\
 & 10 & 17.64 & 29.75 \\
\midrule
NER & 13.6 & 16.33 & 25.24 \\
NCh & 33.2 & 7.86 & 17.75 \\
NER + NCh & 35.4 & 8.97 & 18.84 \\
T5-small & 1 & \bf 37.56 & \bf 49.16 \\
\bottomrule
\end{tabular}
\caption{\textbf{Average over multiple candidates (EM-Avg, F1-Avg).} Measuring what percentage of the proposed answers were also selected as an answer by a human.}
\label{table:2}
\end{table}

\subsubsection{Average Over Multiple Candidates}

Table~\ref{table:2} shows the results for EM-Avg and F1-Avg, i.e.,~measuring what percentage of the proposed answers were also selected as an answer by a human.

Due to the ability of the classifier to take a lower number of candidate questions, we can see that it outperforms taking all named entities or all noun chunks by a sizable margin.

We further see that the average scores consistently drop with the increase of the number of answer candidates. This also explains the lower scores of the named entity and noun chunks approaches as they produce much longer lists of candidate answers.

\subsubsection{Single Answer Candidate}

Finally, we see in both tables, that the T5 model achieves the highest average result. However, in our setup it cannot produce multiple candidates. We plan to extend it accordingly in future work.

\section{Discussion}

\begin{figure}[t!]
\textbf{Context:}
\textit{  \textcolor{brown}{\underline{Oxygen}} is a chemical element with symbol \textcolor{brown}{O} and atomic number \textcolor{brown}{\underline{8}}. It is \textcolor{brown}{a member of the \underline{chalcogen} group} on the periodic table and is a highly reactive nonmetal and oxidizing agent that readily forms compounds (notably oxides) with most elements. By mass, \underline{oxygen is the third-most abundant element} \underline{in the universe, after hydrogen and helium}. At standard temperature and pressure, \textcolor{brown}{\underline{two atoms} of the element} bind to form \underline{dioxygen}, a colorless and odorless diatomic gas with \textcolor{brown}{the formula O2}. \textcolor{brown}{\underline{Diatomic oxygen gas}} constitutes \textcolor{brown}{\underline{20.8\%} of the Earth's atmosphere}. However, monitoring of atmospheric oxygen levels show a global \underline{downward} trend, because of \textcolor{brown}{fossil-fuel burning}. Oxygen is the most abundant element by mass in the Earth's crust as part of \underline{oxide} compounds such as silicon dioxide, making up \textcolor{brown}{\underline{almost half}} of the crust's mass.}
\\ \\ 
\textbf{Top 10 answers:}
\begin{enumerate}
\itemsep0em 
\setlength{\parskip}{0pt}
\setlength{\parsep}{0pt}
\item 8
\item 20.8 \% of the  Earth 's atmosphere
\item Oxygen
\item a member of the chalcogen group 
\item Diatomic oxygen gas
\item almost half
\item O
\item two atoms of the element
\item the formula O2
\item fossil-fuel burning
\end{enumerate}

\caption{The top-10 answer candidates generated by our model for a sample passage from SQuAD1.1. The human-selected ground truth answers are underlined, and the answer candidates are shown in brown.}
\label{fig:x nb example}
\end{figure}

Figure \ref{fig:x nb example} shows a passage from the development split of the SQuAD1.1 dataset and the top-10 answers that our model proposed for it. We can see that these answers represent a diverse set, including named entities, noun chunks, and individual words.
Indeed, this is a typical example, as our analysis across the entire development dataset shows that on average, among the top-10 candidates, our model proposes 4.82 named entities and 6.40 noun chunks. 

Note also that our evaluation setup could be unfair to the model in some cases, e.g.,~if the model proposes a good candidate answer but one that was not chosen by the human annotators, it would receive no credit for it.

Finally, note that our model can produce top-$N$ results for user-defined values of $N$, which is an advantage over simple baselines based on entities or chunks, as well as over our setup for T5.

\section{Conclusion and Future Work}

We proposed a new task: generate answer candidates that can serve as an input to answer-aware question generation models. We further created a dataset for this new task. Moreover, we proposed a suitable model for the task, which combines orthographic, lexical, syntactic, and semantic information, and can generate a pre-specified number of answers. Finally, we demonstrated improvements over simple approaches based on named entities, and competitiveness over complex, computationally expensive neural network models such as T5.

In future work, we plan to analyze and to improve the features. We also want to extend T5 to generate multiple candidates. We further plan to reduce the impact of false negatives, e.g.,~by means of manual evaluation by domain experts, and eventually by producing datasets with (potentially ranked) annotations of all suitable candidate answers.

\section*{Acknowledgments}

This research is partially funded via Project UNITe by the OP ``Science and Education for Smart Growth'' and co-funded by the EU through the ESI Funds under GA No. BG05M2OP001-1.001-0004.

\bibliographystyle{acl_natbib}
\bibliography{ranlp2021}

\end{document}